\documentclass[10pt,twocolumn,letterpaper]{article}

\usepackage[pagenumbers]{cvpr} 

\definecolor{cvprblue}{rgb}{0.21,0.49,0.74}
\usepackage{xcolor}
\definecolor{mypink}{rgb}{1,0.4,0.6}

\usepackage[pagebackref,breaklinks,colorlinks,allcolors=cvprblue,urlcolor=mypink]{hyperref}

\usepackage{amsfonts}
\usepackage{amsmath,amssymb}
\usepackage{booktabs}
\usepackage{multirow}
\usepackage{colortbl}
\usepackage{soul, color, xcolor}
\definecolor{y}{HTML}{FEFAED}
\definecolor{b}{HTML}{EDF9FE}

\def\confName{CVPR}
\def\confYear{2025}

\def\sysName{HiLoTs}

\title{\sysName: High-Low Temporal Sensitive Representation Learning for \\Semi-Supervised LiDAR Segmentation in Autonomous Driving}

\author{R.D. Lin$^1$, Pengcheng Weng$^1$, Yinqiao Wang$^1$, Han Ding$^2$, Jinsong Han$^3$, Fei Wang$^{1\dagger}$\\
 \textit{$^1$ School of Software Engineering, Xi'an Jiaotong University, China} \\
\textit{$^2$ School of Computer Science and Technology, Xi'an Jiaotong University, China}\\
\textit{$^3$ College of Computer Science and Technology, Zhejiang University, China}\\
{\tt\small  rdlin@stu.xjtu.edu.cn,  hanjinsong@zju.edu.cn,  \{dinghan,feynmanw\}@xjtu.edu.cn}\\
{\small $^\dagger$corresponding author and project lead}\\
{\small \url{https://github.com/rdlin118/HiLoTs}}
}

\begin{document}
\maketitle

\begin{abstract}
LiDAR point cloud semantic segmentation plays a crucial role in autonomous driving. In recent years, semi-supervised methods have gained popularity due to their significant reduction in annotation labor and time costs. Current semi-supervised methods typically focus on point cloud spatial distribution or consider short-term temporal representations, e.g., only two adjacent frames, often overlooking the rich long-term temporal properties inherent in autonomous driving scenarios. In driving experience, we observe that nearby objects, such as roads and vehicles, remain stable while driving, whereas distant objects exhibit greater variability in category and shape. This natural phenomenon is also captured by LiDAR, which reflects lower temporal sensitivity for nearby objects and higher sensitivity for distant ones. To leverage these characteristics, we propose \sysName, which learns high-temporal sensitivity and low-temporal sensitivity representations from continuous LiDAR frames. These representations are further enhanced and fused using a cross-attention mechanism. Additionally, we employ a teacher-student framework to align the representations learned by the labeled and unlabeled branches, effectively utilizing the large amounts of unlabeled data. Experimental results on the SemanticKITTI and nuScenes datasets demonstrate that our proposed \sysName~outperforms state-of-the-art semi-supervised methods, and achieves performance close to LiDAR+Camera multimodal approaches. 
\end{abstract}

\section{Introduction}\label{sec:intro}

\begin{figure}[t]
\centering
\includegraphics[width=\columnwidth]{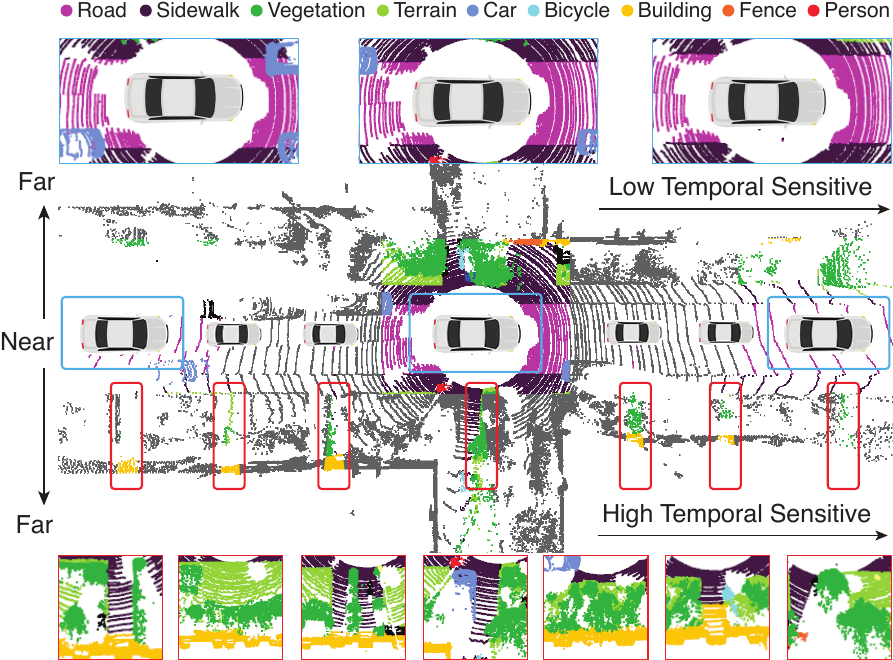}
\caption{Different semantic classes exhibit varying degrees of sensitivity to temporal changes. Objects farther from the vehicle (e.g., vegetation, building, person, etc.) change more frequently over time, as indicated by the red box. In contrast, objects closer to the vehicle (e.g., road, sidewalk, etc.) are less sensitive to temporal changes, as shown by the blue box.}
\label{fig01}
\end{figure}

LiDAR point cloud semantic segmentation is crucial in autonomous driving for tasks such as obstacle avoidance~\cite{c10:zhou2018}, lane detection~\cite{c9:huang2021}, localization and mapping~\cite{c8:jhaldiyal2023, chang2025drivingrulesbenchmarkintegrating}. Most existing segmentation works are fully-supervised approaches~\cite{c32:li2023, c56:fei2021pillarsegnet, c52:zhu2021cylindrical, c40:ando2023}, which have several drawbacks. They require extensive point-wise annotations, leading to significant labor costs and time consumption. Additionally, the need for large labeled datasets limits their scalability and adaptability to new environments, highlighting the need for more efficient and scalable approaches, such as semi-supervised learning~(SSL) methods that can leverage unlabeled data and reduce reliance on costly annotations.

To achieve SSL LiDAR segmentation, many works leverage the spatial distribution information of point clouds~\cite{c12:li2022, c34:cheng2021,c48:li2024}. For example, SSPC-Net~\cite{c34:cheng2021} introduces an area partition method, constructing a super-point graph structure with both labeled and unlabeled point clouds for semi-supervised learning. DDSemi~\cite{c48:li2024} addresses outlier issues in point-to-point SSL methods by employing a point-density-guided contrastive learning technique. In addition to utilizing the spatial distribution of point clouds, some works exploit temporal information~\cite{c21:aygun2021,c22:choy2019, c37:liu2023,c24:shi2022}. For instance, Ayg\"{u}n et al.~ \cite{c21:aygun2021} directly fuses multi-frame point clouds and inputs them into a standard encoder-decoder network to produce segmentation results. Shi et al.~\cite{c24:shi2022} proposes a temporal matching method, performing one-to-one matching based on differences and similarities between point clouds of two consecutive frames. 
Although these methods demonstrate strong performance, they tend to focus on either the spatial or the temporal characteristics of point clouds, without fully integrating both aspects.

In driving experience, we observe a phenomenon: objects closer to the vehicle, such as roads and cars, tend to have stable categories and shapes as the vehicle moves, while distant objects, such as pedestrians, guardrails, plants, and buildings, exhibit significant variations in category and shape. Surprisingly,  this nature is also reflected in LiDAR point cloud data, as shown in Fig.~\ref{fig01}, where the relevant areas are highlighted. To leverage this phenomenon, we propose \sysName, which consists of a High Temporal Sensitivity Flow (HTSF) and a Low Temporal Sensitivity Flow (LTSF). The HTSF focuses on regions where distant objects experience significant changes in category and shape, while the LTSF focuses on nearby regions where object categories and shapes remain relatively stable. Furthermore, the features from HTSF and LTSF are fused and interact through a cross-attention mechanism. To better represent near and far objects, we convert the point cloud into cylindrical voxels using a cylindrical voxelization network~\cite{c52:zhu2021cylindrical}. To further optimize computational efficiency, we aggregate multiple spatiotemporally neighboring cylindrical voxels, enabling a more efficient computation of HTSF and LTSF.

For semi-supervised point cloud segmentation, we adopt the mainstream Mean Teacher architecture \cite{c14:tarvainen2017}, which effectively leverages a small amount of labeled LiDAR frames and a large amount of unlabeled data. In \sysName, the labeled LiDAR frames are fed into the student network, while the unlabeled LiDAR frames are processed through the teacher network. In each iteration, a consistency loss is computed to align the predictions made by the teacher network with those from the student network. The student network gradually updates the teacher network’s parameters, a process that can be likened to the student slowly growing into the teacher. After training, the teacher network is used for the LiDAR segmentation during inference.

We evaluate \sysName~on two widely-used autonomous driving benchmarks, SemanticKITTI and nuScenes. Extensive results show that \sysName~outperforms the latest LiDAR-only semi-supervised methods and achieves performance comparable to multimodal approaches such as~\cite{c46:chen2024, c53:kong2024multi}, which combine LiDAR and camera data. Additionally, ablation studies confirm the effectiveness of the proposed HTSF and LTSF components, aligning with our observations from driving experience. 
In summary, our work claims the following main contributions:

\begin{itemize}
    \item We observe a natural but often overlooked phenomenon in driving as shown in Fig.~\ref{fig01}. We propose \sysName, designed to focus on this characteristic. We believe this design could provide valuable insight for future advancements in LiDAR segmentation tasks.

    \item  \sysName~includes several novel techniques, such as multi-voxel aggregation and temporal sensitivity embedding units, which are efficient and effective in LiDAR spatiotemporal representation learning. 
    
    \item Experimental results show that \sysName~surpasses the latest semi-supervised methods and achieves performance comparable to LiDAR+Camera multimodal methods. 
\end{itemize}

\section{Related Work}\label{sec:relatedwork}

\begin{figure*}[t]
\centering
\includegraphics[width=0.95\textwidth]{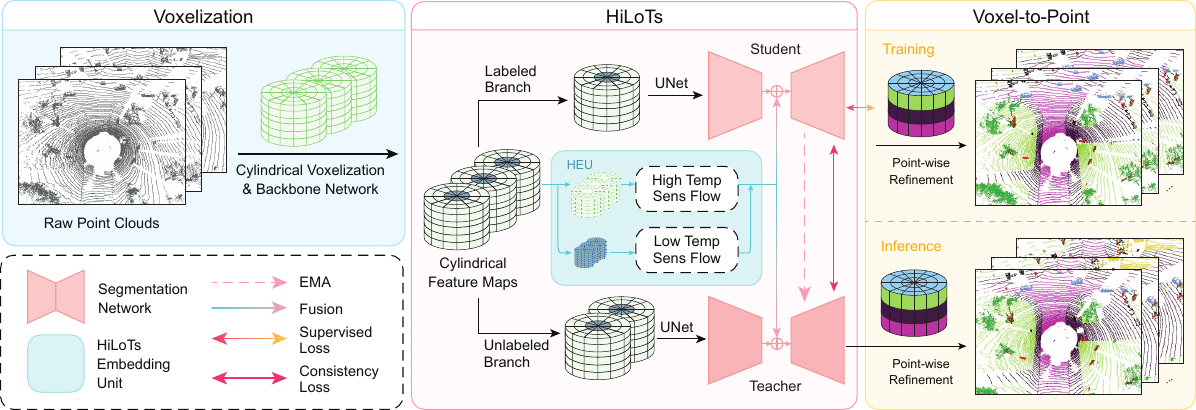}

\caption{Our segmentation model involves three stages. During voxelization, cylindrical voxelization is applied to transform unordered points into volumetric grids, followed by a spatial feature extraction backbone. Then, \sysName~processes the labeled and unlabeled cylindrical features through a student-teacher framework. It also integrates the attention map from \sysName~embedding unit~(HEU) to produce voxel-level segmentation maps. 
Finally, a point-wise refinement network is utilized to obtain point-level segmentation results. }
\label{fig02}
\end{figure*}

\subsection{Semi-supervised LiDAR Segmentation}

Various works utilize LiDAR to capture objects' 3D representation since point clouds can accurately reflect the structural characteristics of objects \cite{c28:li2023, c29:thomas2019, c30:lang2019, c31:lai2023spherical, c32:li2023}. However, most of the existing works are based on fully-supervised learning, while it is difficult to annotate point clouds due to the intensive labor and time costs. Consequently, many works have attempted to leverage SSL methods \cite{c13:jiang2021, c17:kong2023, c33:li2023, c34:cheng2021, c35:xu2023, c47:unal2024,c48:li2024}. GPC \cite{c13:jiang2021} uses large amounts of unlabeled data as pseudo-label guidance to reduce the negative impact of intra-class negative pairs, achieving good results in both indoor and outdoor scenarios. LaserMix \cite{c17:kong2023} deeply analyzes point cloud distribution priors of various objects and performed fusion between labeled and unlabeled point clouds.
Nevertheless, continuous point clouds in autonomous driving usually contain rich temporal features. Current SSL methods mainly focus on spatial feature extraction, neglecting the inherent temporal representations of outdoor point clouds.

\subsection{Point Cloud Representation }

Mainstream outdoor point cloud representation learning can be analyzed from both spatial and temporal perspectives. For spatial representation, since 3D point clouds are unordered sets \cite{c38:qi2017}, it is necessary to first convert the point clouds into volumetric grids, where it can be input into neural networks as tensors. Current mainstream approaches include range mapping \cite{c40:ando2023, c54:kong2023rethinking}, cubic-voxelization \cite{c55:ye2023lidarmultinet}, pillar-based voxelization \cite{c30:lang2019, c28:li2023}, spherical \cite{c31:lai2023spherical} and cylindrical representation \cite{c52:zhu2021cylindrical}. For temporal representation, commonly used methods can be divided into data-level~\cite{c21:aygun2021, c22:choy2019, c37:liu2023} and feature-level~\cite{c24:shi2022} multi-frame fusion. 
However, data-level fusion~\cite{c21:aygun2021, c22:choy2019} directly increases computational overhead, making hardware resources a primary bottleneck. Also, current feature-level fusion in semi-supervised learning~\cite{c24:shi2022} is based on two adjacent frames, which is insufficient to encode temporal changes compared to multiple frames. We observe that during driving, distant objects show frequent changes in both object categories and shapes over time, while closer objects exhibit more stable distribution. To fully capture this property, we propose \sysName.

\section{Methods}\label{sec:methods}
\label{sec:method}

Fig.~\ref{fig02} illustrates the overall pipeline of the proposed method. We first introduce the preliminary of the task. Next, we provide a detailed description of each component in the proposed \sysName. Finally, we present the loss functions for training the overall network. 

\textbf{Preliminary.} 
Suppose $D = \{D_i \;|\; i = 1, \cdots, N\}$ denotes the point cloud dataset, where $N$ is the total number of sequences. Each sequence $D_i$ includes $t_i$ point cloud frames, $ \{f_j \;|\; j = 1, \cdots, t_i\}$. $f_{j} = (x_p, y_p, z_p, r_p) \in \mathbb{R}^{P\times 4}$ represents $P$ point clouds in one frame, where $(x_p, y_p, z_p)$  and $r_p$ denote the Cartesian coordinates and LiDAR intensity, respectively. In the semi-supervised semantic segmentation task, given a supervised ratio of $s\%$, $D_{i} = [f_1, \cdots, f_{t_i}]$ contains uniformly sampled  $s\%$ labeled frames, and $(1-s)\%$ unlabeled frames.

\subsection{Voxelization with Cylindrical Network}\label{sec:point-to-voxel}

Common visual modalities such as images and videos are in volumetric grids, which are efficient for neural network processing. However, point clouds are unordered sets \cite{c38:qi2017}. To transform point clouds to volumetric grids for network processing, we employ a mainstream and efficient feature extraction method, namely cylindrical voxelization~\cite{c52:zhu2021cylindrical}, which can adequately represent point clouds of different densities at different ranges.

The first step of cylindrical voxelization is to convert the Cartesian coordinates of each point, represented as $(x, y, z)$, into the corresponding cylindrical coordinates $(\rho, \theta, z)$, where $\rho = \sqrt{x^2+y^2}$ and $\theta = \mathrm{arctan}(\frac{y}{x})$ represent the radial distance and the azimuth, respectively.

Next, since LiDAR captures dense point clouds in near areas and sparse point clouds in far areas, during voxelization, we set the size of cylindrical cells to increase with the distance, maintaining a balanced number of points in different cylindrical cells. Each cell contains the cylindrical coordinate $(\rho, \theta, z)$ of the original point cloud, with the remission $r_\epsilon$. For multiple points projected onto the same cell, we retain the point with the closest range. The resulting volumetric grid of each frame is  $x_f \in \mathbb{R}^{\mathrm{R} \times \Theta \times H \times C}$, where $C=4$ containing $(\rho, \theta, z, r_{\epsilon})$, and $\mathrm{R}$, $\Theta$ and $H$ represent the maximum radius, azimuth, and height, respectively.

Further, we employ 3D ResNet50~\cite{c41:he2016}  to extract the initial cylindrical features for the following LiDAR segmentation, $x_f \to  x_f \in \mathbb{R}^{M\times d}$, where $M=\mathrm{R} \times \Theta \times H$, $d=256$.

\subsection{HiLoTs Embedding Unit}\label{sec:heu}

\begin{figure}[t]
\centering
\includegraphics[width=\columnwidth]{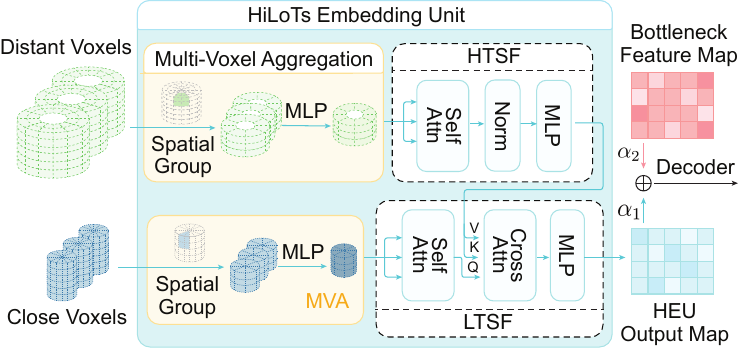}
\caption{\sysName~Embedding Unit (HEU). The distant voxel features are passed into high temporal sensitivity flow, while voxel features in closer areas undergo low temporal sensitivity flow. The output map of HEU is fused with the bottleneck feature map from the segmentation model, further passed into the decoder.}
\label{fig03}
\end{figure}

We observe that nearby objects, mainly roads and vehicles, remain relatively stable in their spatial and categorical characteristics during driving, while distant objects display greater variability in both category and shape. Leveraging this observation, we propose \sysName~Embedding Unit~(HEU) that distinguishes between high temporally sensitive features for distant scenes and low temporally sensitive features for nearby scenes. By capturing these distinctions, \sysName~can more accurately interpret dynamic scenes and better adapt to the varying spatial-temporal characteristics across different distances.

As shown in Fig.~\ref{fig03}, HEU processes two types of 
initial cylindrical features, which first undergo Multi-Voxel Aggregation. Voxels located at distant ranges are directed through the High Temporal Sensitivity Flow, while those within closer ranges are processed via the Low Temporal Sensitivity Flow. The resulting feature maps from each flow are then integrated, which facilitates interaction between the high- and low-sensitivity features. These enriched features subsequently serve as inputs to the segmentation network.

\subsubsection{Multi-Voxel Aggregation}\label{sec:multi-voxel-aggregation}

Recent Transformer-based models~\cite{c25:vaswani2017, c42:dosovitskiy2021, c43:yan2024, lan2024bullydetect, shi2023exploiting} have demonstrated remarkable performance in various tasks such as LiDAR semantic segmentation~\cite{c40:ando2023, c31:lai2023spherical}. Given that Transformers serve as a unifying foundation for multimodal modeling, we also employ a Transformer-based architecture to implement our HEU model. However, the computational complexity of the attention mechanism is $O(n^2)$, where $n$ represents the number of tokens. With the extensive number of voxels in outdoor scenes, applying the attention mechanism to all voxels becomes computationally prohibitive. To address this, we designed a multi-voxel aggregation method~(MVA), which groups neighboring voxels into super-voxels. This approach not only significantly reduces the token count, thus alleviating computational demands, but also aggregates voxel features, enhancing the model’s performance by capturing more coherent semantic information within each super-voxel. As shown in Fig. \ref{fig03}, the proposed MVA includes two steps. \textbf{(1) Spatial aggregation:} we aggregate $M$ cylindrical voxels into $m$ super voxels in spatial as:
\begin{equation}\label{eqn:mva-spatial}
F' = \mathrm{MLP}_{\theta_1}(\mathrm{NNGroup}(F, m))
\end{equation}
where $F\in\mathbb{R}^{M\times d\times t}$ represents cylindrical features across $t$ frames; $\mathrm{NNGroup}(\cdot)$ represents nearest neighbor grouping and $\mathrm{MLP}_{\theta_1}$ is a multi-layer perceptron (MLP) with parameters of ${\theta_1}$. $F'\in \mathbb{R}^{m\times d\times t}$ ($m<M$) is the spatially aggregated cylindrical features.

\textbf{(2) Temporal fusion:} we further fuse super-voxel features across temporal dimension as:
\begin{equation}\label{eqn:mva-temporal}
    V = \mathrm{MLP}_{\theta_2}(\mathrm{AvgPool}(F') )
\end{equation}
where $\mathrm{AvgPool}(\cdot)$ represents temporal average pooling and $\mathrm{MLP}_{\theta_2}$ is an MLP with parameters of ${\theta_2}$. $V\in \mathbb{R}^{m\times d}$ is the final output of the Multi-Voxel Aggregation process.

\subsubsection{High Temporal Sensitivity Flow}

The High Temporal Sensitivity Flow (HTSF) is designed to process aggregated voxels at greater distances, e.g., those within the farthest 70\% of the range. This flow enables point cloud representation learning by focusing on temporally dynamic and spatially variable distant regions, where object categories and shapes are more likely to fluctuate. The process can be represented as follows:
\begin{equation}\label{eqn:4}
\begin{aligned}
        V_{i} = \mathrm{Softmax}&\left(\frac{\text{dot}(V_i W_Q, V_i W_K)}{\sqrt{d_k}}\right) V_i W_i \\
        &V_{i+1} = \mathrm{MLP}_i(V_{i})
\end{aligned}
\end{equation}
where $V_i \in \mathbb{R}^{m\times d}$ represents the far-range voxel features of the $i$-th encoder layer, $W_Q \in \mathbb{R}^{d\times d_k}$, $W_K \in \mathbb{R}^{d\times d_k}$ and $W_V \in \mathbb{R}^{d\times d_v}$ denotes the weight parameters for Query, Key and Value matrices. $\text{dot}(\cdot, \cdot)$ denotes matrix multiplication. $V_{i+1} \in \mathbb{R}^{m\times d}$ represents the output of the current encoder layer, which is also the input to the next layer.

\subsubsection{Low Temporal Sensitivity Flow}

As discussed, the point cloud distribution of closer objects, such as roads, remains relatively stable over time in LiDAR point cloud. To capture these low temporally sensitive features accurately, we design the Low Temporal Sensitivity Flow (LTSF), depicted in the lower part of Fig.~\ref{fig03}. In this process, all distant voxels are first excluded from the voxel set, leaving only those that represent low temporal sensitivity voxel features. These selected voxels then undergo Multi-Voxel Aggregation and self-attention, similar to the HTSF process. Further, a cross-attention is applied with the attention map generated by HTSF, enabling an interaction between the high and low temporal sensitivity features. 

By stacking multiple HTSF and LTSF layers, we obtain the final output of the \sysName~ Embedding Unit. To fuse the HEU output with the bottleneck encoder feature map of the segmentation network, we introduce two learnable parameters, $\alpha_1$ and $\alpha_2$, which control the feature fusion process.
\begin{equation}\label{eqn:7}
    S = \alpha_1 \cdot \mathrm{BottleNeck_{En}}(x_f) + \alpha_2 \cdot \mathrm{HEU}(F)
\end{equation}
where $x_f$ and $F$ are described in Sec.~\ref{sec:point-to-voxel} and 
Equation.~\ref{eqn:mva-spatial}. 
Finally, the decoder of the bottleneck network takes $S$ as input and outputs the voxel-level features, denoted as $\mathrm{BottleNeck_{De}}(S) \to S \in \mathbb{R}^{M\times d}$. We use  MinkowskiUNet~\cite{c22:choy2019} as the bottleneck network for segmentation.

\subsection{Voxel to Point Cloud Results}\label{sec:voxel-to-point}

Since \sysName~employs a voxelized approach to convert point clouds into volumetric grids, this inevitably leads to information loss when points from different objects are mapped to the same voxel. We first reverse voxel-level features $S$ to point-level features, $S \to \mathbb{R}^{P\times d}$, where $P$ represents the number of points in the frame, based on the point-to-voxel mapping table from voxelization process described in Sec.~\ref{sec:point-to-voxel}. Then, we apply a point-wise refinement network~\cite{c52:zhu2021cylindrical} to output point-level segmentation results, where $S$ are fused with the point coordinates and LiDAR intensity.
\begin{equation}\label{eq:refinement}
    \hat{y} = \mathrm{RefineNet} (S, (x_p,y_p,z_p,r_p))
\end{equation}
where $\hat{y} \in \mathbb{R}^{P\times K}$ represents $K$ object classification confidences for $P$ points, i.e., semantic segmentation results.

\subsection{Semi-supervised Learning and Loss Functions}\label{sec:loss-function}

Our \sysName~model adopts Mean Teacher architecture \cite{c14:tarvainen2017} to achieve semi-supervised segmentation of point clouds. It includes two segmentation networks, namely the student and teacher network. The student network receives the labeled point clouds and uses the corresponding ground-truth for supervised training. Specifically, we use focal loss \cite{c51:lin2017} as the supervised loss function, which addresses the class imbalance problem in point cloud semantic segmentation.
\begin{equation}\label{eqn:8}
    L_{sup} = \mathrm{FocalLoss}(\hat{y}_s, y)
\end{equation}
where $\hat{y}_s$ represents the student network's segmentation prediction and $y$ represents the ground-truth labels.

In contrast, the teacher network receives unlabeled data, which are also fed into the student network. The teacher network’s loss function is the consistency loss between the two networks, represented as:
\begin{equation}\label{eqn:9}
    L_{con} = \lVert \hat{y}_s - \hat{y}_t \rVert _2
\end{equation}
where $\hat{y}_t$ denotes the prediction from the teacher network; $\lVert \cdot \rVert _2$ represents the $\mathcal{L}_2$ norm. 
After obtaining the supervised loss and consistency loss, the final loss of the model is the weighted sum of the two:
\begin{equation}\label{eqn:10}
    L = \alpha L_{{sup}} + \beta L_{{con}}
\end{equation}
where $\alpha$ and $\beta$ are to balance two losses. In our experiments, we set both to 1.

The weights of the teacher network are initialized by exponential moving average (EMA)~\cite{c14:tarvainen2017}, transferring the student network's trained parameters to the teacher model, which can be represented as follows:
\begin{equation}\label{eq:ema}
    W'_t = \gamma W'_{t-1} + (1-\gamma)W_t
\end{equation}
where $W'_t$ and $W$ are for teacher network and student network at the time of $t$, respectively. The process of mean-teacher semi-supervised learning is akin to a student gradually growing into a teacher, slowly absorbing knowledge and refining their understanding over time. Ultimately, we obtain the teacher network, which is then used for inference.

\subsection{Implementation Details}

\sysName~is trained for 50,000 iterations with an early stopping strategy. AdamW optimizer~\cite{c49:loshchilov2018} is used with the initial learning rate set to 1e-3. All experiments are conducted on a server with four RTX 3090 GPUs, using a batch size of 4 per GPU, resulting in a total batch size of 16. In cylindrical voxelization, we set the maximum point cloud range, azimuth, and height to $(0, 50)$ meters,  $(-\pi, \pi)$, and $(-4, 2)$ meters, respectively. The resolution of voxelized grids is set to (240, 180, 20). Considering both the model performance and the GPU memory cost, we set $t=5$ as the temporal length in all experiments. The layers of both encoder and decoder of the Transformer are set to $N=6$.

During training, \sysName~takes 1 frame (central frame) and its neighboring $t-1$ frames as input. If the central frame is labeled, it is processed by the labeled branch for supervised segmentation. Otherwise, it is routed to the unlabeled branch. Regardless of label presence, all $t$ frames pass through the HEU module. During inference, \sysName~takes $t$-frame LiDAR point clouds as input and leverages the teacher network to estimate semantic segmentation for the central frame.

\begin{table*}[t]
\small
\centering
\renewcommand\arraystretch{1.0}
\begin{tabular}{lccccccccc}
\toprule
\multirow{2}{*}{\hspace{2pt}Methods} & \multirow{2}{*}{Modality} & \multicolumn{4}{c}{SemanticKITTI} & \multicolumn{4}{c}{nuScenes}\\
\cmidrule(l){3-10}
& & \makebox[2.5em][c]{1\%}  & \makebox[2.5em][c]{10\%}  & \makebox[2.5em][c]{20\%}  & \makebox[2.5em][c]{50\%} & \makebox[2.5em][c]{1\%}  & \makebox[2.5em][c]{10\%}  & \makebox[2.5em][c]{20\%} & \makebox[2.5em][c]{50\%}\\
\midrule
\cellcolor{y} Cylinder3D \cite{c52:zhu2021cylindrical} (2021) & Lidar & 45.4 & 56.1 & 57.8 & 58.7 & 53.4$^{\ast}$ & 63.4$^{\ast}$ & 67.0$^{\ast}$ & 71.9$^{\ast}$\\
\cellcolor{y} RangeViT \cite{c40:ando2023} (2023) & Lidar & 43.8$^{\ast}$ & 53.4$^{\ast}$ & 56.6$^{\ast}$ & 58.8$^{\ast}$ & 53.8$^{\ast}$ & 64.6$^{\ast}$ & 67.8$^{\ast}$ & 73.1$^{\ast}$\\
\cellcolor{y} SphereFormer \cite{c31:lai2023spherical} (2023) & Lidar & 41.2$^{\ast}$ & 59.8$^{\ast}$ & 60.6$^{\ast}$ & 62.4$^{\ast}$ & 49.5$^{\ast}$ & 65.3$^{\ast}$ & 69.2$^{\ast}$ & 73.7$^{\ast}$\\
\cellcolor{y} MarS3D \cite{c37:liu2023} (2023) & Lidar & 44.5$^{\ast}$ & 58.6$^{\ast}$ & 60.2$^{\ast}$ & 61.7$^{\ast}$ & 51.8$^{\ast}$ & 65.5$^{\ast}$ & 68.4$^{\ast}$ & 72.8$^{\ast}$\\
\midrule
\cellcolor{b} GPC \cite{c13:jiang2021} (2021) & Lidar & 41.8 & 49.9 & 58.8  & 59.9 & - & - & - & -\\
\cellcolor{b} PolarMix \cite{xiao2022polarmix} (2022) & Lidar & 50.1 & 60.9 & 62.0 & 63.8 & 55.6 & 69.6 & 71.0 & 73.8\\
\cellcolor{b} LaserMix \cite{c17:kong2023} (2023) & Lidar & 50.6 & 60.0 & 61.9 & 62.3 & 55.3 & 69.9 & 71.8 & 73.2\\
\cellcolor{b} LiM3D \cite{c33:li2023} (2023) & Lidar & 58.4 & 62.2 & 63.1 & 63.6 & -  & - & - & -\\
\cellcolor{b} ImageTo360 \cite{reichardt2023360deg} (2023) & Lidar & 54.5 & 58.6 & 61.4 & 64.2 & - & - & - & -\\
\cellcolor{b} IGNet \cite{c47:unal2024} (2024) & Lidar & 49.0 & 61.3 & 63.1 & 64.8 & - & - & - & - \\
\cellcolor{b} DDSemi \cite{c48:li2024} (2024) & Lidar & \textbf{59.3} & \underline{65.1}  & \underline{66.3} & \underline{67.0} & 58.1 & 70.2 & 74.0 & \underline{76.5}\\
\cellcolor{b} FRNet \cite{xu2025frnet} (2025) & Lidar & 55.8 & 64.8 & 65.2 & 65.4 & \textbf{61.2} & \textbf{72.2} & \underline{74.6} & 75.4\\
\hline
\cellcolor{b} CyMix+IPSL \cite{c46:chen2024} (2024) & Lidar+Camera & \textit{52.8} & \textit{64.8} & \textit{64.9} & \textit{65.9} & \textit{59.1} & \textit{76.0} & \textit{78.7} & \textit{80.5}\\
\cellcolor{b} LaserMix++ \cite{c53:kong2024multi} (2024) & Lidar+Camera & \textit{63.2} & \textit{67.5} & \textit{67.7} & \textit{68.6} & \textit{65.3} & \textit{75.3} & \textit{75.2} & \textit{76.3} \\
\midrule
\hspace{0.2pt} \sysName~(Ours) & Lidar & \underline{58.6} & \textbf{65.7} & \textbf{66.5} & \textbf{67.6} & \underline{58.7} & \textbf{72.2} & \textbf{75.2} & \textbf{76.9} \\
\bottomrule
\end{tabular}
\caption{Performance comparisons with current state-of-the-art. Methods highlighted in \sethlcolor{y}\hl{yellow} represents fully-supervised methods, while \sethlcolor{b}\hl{blue} denotes semi-supervised methods. Best scores are \textbf{bolded} and the second best scores are \underline{underlined}. Results in ${\ast}$ are reproduced. }
\label{table01}
\end{table*}

\section{Experiments and Analysis}\label{sec:experiments}

\subsection{Datasets and Evaluation Metric}

We evaluate our approach on two widely-used autonomous driving datasets: SemanticKITTI~\cite{c26:behley2019} and nuScenes~\cite{c27:hulger2020}. LiDAR point clouds from SemanticKITTI were collected with a 64-beam Velodyne HDL-64E device at 10Hz, including 19 semantic classes. Sequences 00-07 and 09-10 serve as the training set, with 19120 point cloud scenes, while sequence 08 is used as the validation set, with 4070 scenes. The nuScenes dataset, also widely used, was collected with a 32-beam Velodyne HDL-32E device at 20Hz. It contains 16 semantic classes, with the training and validation sets of 27287 and 5850 point cloud scenes, respectively. 

Following prior semi-supervised segmentation works~\cite{c17:kong2023, c33:li2023, c48:li2024}, we report the model's performance on the validation sets of both datasets. We set the labeled ratio in \{$1\%, 10\%, 20\%, 50\%$\} and use mean Intersection-over-Union (mIoU) scores across all semantic classes as the evaluation metric.

\subsection{Performance Comparisons}

Table~\ref{table01} presents a comparison of \sysName~with supervised methods with limited labeled data~\cite{c52:zhu2021cylindrical, c40:ando2023, c31:lai2023spherical, c37:liu2023}, LiDAR-only semi-supervised methods~\cite{c13:jiang2021, xiao2022polarmix, c33:li2023, reichardt2023360deg, c47:unal2024, c48:li2024, xu2025frnet}, and LiDAR+Camera semi-supervised methods~\cite{c46:chen2024, c53:kong2024multi}.

\noindent\textbf{Comparisons with Fully-supervised Methods. }
Cylinder3D~\cite{c52:zhu2021cylindrical}, RangeViT~\cite{c40:ando2023}, SphereFormer~\cite{c31:lai2023spherical}, and MarS3D~\cite{c37:liu2023} are fully supervised methods that do not incorporate semi-supervised techniques. We re-train both methods using the same amount of labeled data as our semi-supervised approach, with re-trained results marked with the $^*$ in Table~\ref{table01}. \sysName~ consistently outperforms these fully supervised approaches by a notable margin. For instance, using only 1\% SemanticKITTI labeled dataset, our method achieves a mIoU of 58.6, while Cylinder3D and MarS3D reach 45.4 and 44.5, respectively. This is because the mean-teacher semi-supervised strategy can effectively leverage unlabeled data to improve segmentation performance. Although involving more labeled data can gradually enhance the performance of fully supervised methods, the high labeling cost in customized point cloud segmentation tasks makes semi-supervised methods more feasible.

\begin{table*}[t]
\centering
\small
\begin{tabular}{lp{0.35cm}p{0.35cm}p{0.35cm}p{0.35cm}p{0.35cm}p{0.35cm}p{0.35cm}p{0.35cm}p{0.35cm}p{0.35cm}p{0.35cm}p{0.35cm}p{0.35cm}p{0.35cm}p{0.35cm}p{0.35cm}p{0.35cm}p{0.35cm}p{0.35cm}}
\hline
Methods & \rotatebox{55.62}{car} & \rotatebox{55.62}{bicycle} & \rotatebox{55.62}{motorcycle} & \rotatebox{55.62}{truck} & \rotatebox{55.62}{bus} & \rotatebox{55.62}{person} & \rotatebox{55.62}{bicyclist} & \rotatebox{55.62}{road} & \rotatebox{55.62}{parking} & \rotatebox{55.62}{sidewalk} & \rotatebox{55.62}{other-ground} & \rotatebox{55.62}{building} & \rotatebox{55.62}{fence} & \rotatebox{55.62}{vegetation} & \rotatebox{55.62}{trunk} & \rotatebox{55.62}{terrain} & \rotatebox{55.62}{pole} & \rotatebox{55.62}{traffic-sign} & \rotatebox{55.62}{mIoU} \\
\midrule
Cylinder3D & 94.1 & \textbf{38.8} & 43.9 & 42.1 & 38.2 & 57.3 & 72.0 & 91.4 & 38.1 & 76.4 & 0.4 & 89.7 & 56.0 & 86.4 & 59.7 & 70.6 & 60.5 & 45.5 & 55.9 \\
RangeViT & 92.0 & 29.4 & 37.0 & 52.6 & 33.6 & 43.5 & 60.4 & 93.9 & 41.5 & 79.9 & 1.4 & 83.8 & 52.1 & 84.3 & 57.7 & 72.4 & 57.2 & 39.7 & 53.4 \\
GPC & 94.1 & 1.4 & 53.1 & 47.2 & 28.1 & 43.2 & 0.0 & 93.1 & 28.6 & 77.4 & 0.1 & 87.6 & 39.4 & 87.8 & 56.4 & \textbf{77.8} & 60.8 & 50.6 & 48.8 \\
LaserMix & 95.8 & 13.5 & 52.1 & \textbf{71.4} & 50.2 & 66.4 & 55.9 & 93.6 & 48.9 & 81.8 & 0.4 & 91.6 & 66.1 & 88.6 & 66.6 & 75.4 & 65.0 & 49.5 & 59.7 \\
\midrule
HiLoTs & \textbf{96.1} & 33.1 & \textbf{68.1} & 63.0  & \textbf{63.4} & \textbf{71.8} & \textbf{81.5} & \textbf{94.7} & \textbf{60.2} & \textbf{83.9} & \textbf{13.1} & \textbf{92.2} & \textbf{70.1} & \textbf{89.1} & \textbf{70.4} & 76.5 & \textbf{67.0} & \textbf{52.3} & \textbf{65.7} \\
\bottomrule
\end{tabular}
\caption{Class-wise IoU score on the validation set of SemanticKITTI under 10\% supervised ratio. Note that the class `motorcycle' is omitted due to its low distribution in the validation set.}
\label{tab:02}
\end{table*}

\begin{figure*}[t]
\centering
\includegraphics[width=0.915\textwidth]{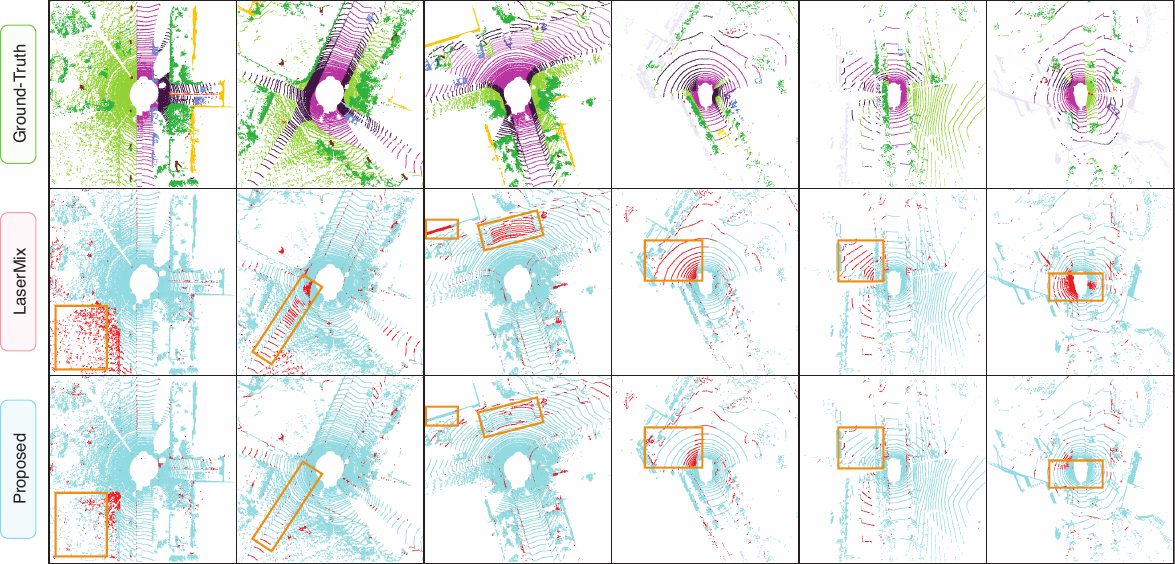}
\caption{Error maps visualization (blue and red points are for correct predictions and incorrect predictions, respectively.). The left three columns are segmentation results from SemanticKITTI dataset, while the right three columns are from nuScenes. Our \sysName~method shows a significant improvement in the area of distant objects. }
\label{fig04}
\end{figure*}

\noindent\textbf{Comparisons with Semi-upervised Methods. }
In mainstream semi-supervised point cloud segmentation for autonomous driving, existing methods can be categorized into single-modal methods that rely solely on LiDAR, and multi-modal methods that incorporate both LiDAR and camera.

As shown in Table~\ref{table01}, \sysName~outperforms most current state-of-the-art LiDAR-only semi-supervised approaches, including recent methods such as FRNet~\cite{xu2025frnet} and DDSemi~\cite{c48:li2024}, across various labeling ratios. This demonstrates the effectiveness of the \sysName~Embedding Unit module, described in Sec.~\ref{sec:heu}, which leverages the distinct temporal and spatial variations of point clouds at different distances, making it highly effective for point cloud segmentation in autonomous driving.

We also compare our method with two recent LiDAR + Camera multimodal semi-supervised approaches, i.e., CyMix+IPSL~\cite{c46:chen2024} and LaserMix++~\cite{c53:kong2024multi}. \sysName~achieves performance comparable to LaserMix++ on both datasets, performing slightly better than CyMix+IPSL on SemanticKITTI dataset and slightly worse on nuScenes dataset. Overall, \sysName~demonstrates considerable effectiveness even when compared to multimodal methods. Its advantage lies in not requiring a camera or labeled RGB data from the camera, making the system more cost-effective and reducing the labeling cost.

\begin{table*}
\caption{Ablation study on the core components of \sysName. }
\small
\centering
\renewcommand\arraystretch{1.0}
\begin{subtable}[t]{0.32\textwidth}
\setlength{\tabcolsep}{2pt}
\begin{tabular}{lcccccc}
\hline
\multirow{2}{*}{Structure} & \multicolumn{3}{c}{SemanticKITTI} & \multicolumn{3}{c}{nuScenes} \\ \cmidrule(l){2-7}
& \makebox[1.5em][c]{10\%} & \makebox[1.5em][c]{20\%}  & \makebox[1.5em][c]{50\%} & \makebox[1.5em][c]{10\%}  & \makebox[1.5em][c]{20\%} & \makebox[1.5em][c]{50\%}  \\ 
\hline
None & 59.2 & 60.3 & 60.9 & 66.8 & 68.4 & 69.2 \\
HTSF & \underline{63.4} & \underline{63.9} & \underline{64.5} & \underline{69.2} & \underline{71.8} & \underline{74.3} \\
LTSF & 62.8 & 63.5 & 64.3 & 68.5 & 71.2 & 73.9 \\
HEU & \textbf{65.7} & \textbf{66.5} & \textbf{67.6} & \textbf{72.2} & \textbf{75.2} & \textbf{76.9} \\
\hline
\end{tabular}
\caption{\textbf{HEU components}. ``None'' denotes no HEU is applied. ``HTSF'' and ``LTSF'' denote only applying high and low temporal sensitivity flow, respectively. ``HEU'' represents the original method. }
\label{table:ablation-heu}
\end{subtable}
\hspace{\fill}
\begin{subtable}[t]{0.32\textwidth}
\setlength{\tabcolsep}{2pt}
\begin{tabular}{lcccccc}
\hline
\multirow{2}{*}{Fusion} & \multicolumn{3}{c}{SemanticKITTI} & \multicolumn{3}{c}{nuScenes} \\ \cmidrule(l){2-7}
& \makebox[1.5em][c]{10\%} & \makebox[1.5em][c]{20\%}  & \makebox[1.5em][c]{50\%} & \makebox[1.5em][c]{10\%}  & \makebox[1.5em][c]{20\%} & \makebox[1.5em][c]{50\%}  \\ 
\hline
Add & 63.6 & 64.7 & 65.3 & 69.8 & 72.8 & 74.6 \\
Concat & 64.3 & 65.1 & 65.9 & 70.3 & 73.5 & 75.2 \\
High as Q & \underline{65.4} & \underline{66.1} & \underline{66.9} & \underline{71.5} & \textbf{75.4} & \underline{76.5} \\
Low as Q & \textbf{65.7} & \textbf{66.5} & \textbf{67.6} & \textbf{72.2} & \underline{75.2} & \textbf{76.9} \\
\hline
\end{tabular}
\caption{\textbf{Fusion in HEU.} ``Add'' and ``Concate'' denote element-wise addition and concatenation, respectively.  ``High as Q'' represents the self-attention map from HTSF serves as Query, and ``Low as Q'' denotes the original method. }
\label{table:ablation-fusion}
\end{subtable}
\hspace{\fill}
\begin{subtable}[t]{0.32\textwidth}
\setlength{\tabcolsep}{2pt}
\begin{tabular}{lcccccc}
\hline
\multirow{2}{*}{Sampling} & \multicolumn{3}{c}{SemanticKITTI} & \multicolumn{3}{c}{nuScenes} \\ \cmidrule(l){2-7}
& \makebox[1.5em][c]{10\%} & \makebox[1.5em][c]{20\%}  & \makebox[1.5em][c]{50\%} & \makebox[1.5em][c]{10\%}  & \makebox[1.5em][c]{20\%} & \makebox[1.5em][c]{50\%}  \\ 
\hline
Random & 63.5 & 64.2 & 64.9 & 69.3 & 72.4 & 73.5 \\
Density & 63.9 & 64.8 & 65.4 & 69.5 & 73.2 & 74.1 \\
Aggregate & \textbf{65.7} & \textbf{66.5} & \textbf{67.6} & \textbf{72.2} & \textbf{75.2} & \textbf{76.9} \\
\hline
\end{tabular}
\caption{\textbf{Voxel down-sampling strategies in MVA}. ``Random'' denotes random selection, and ``Density'' denotes voxels with the most point cloud density. ``Aggregate'' denotes our approach.}
\label{table:ablation-MVA}
\end{subtable}
\begin{subtable}[t]{0.32\textwidth}
\setlength{\tabcolsep}{2pt}
\begin{tabular}{lcccccc}
\hline
\multirow{2}{*}{Backbone} & \multicolumn{3}{c}{SemanticKITTI} & \multicolumn{3}{c}{nuScenes} \\ \cmidrule(l){2-7}
& \makebox[1.5em][c]{10\%} & \makebox[1.5em][c]{20\%}  & \makebox[1.5em][c]{50\%} & \makebox[1.5em][c]{10\%}  & \makebox[1.5em][c]{20\%} & \makebox[1.5em][c]{50\%}  \\ 
\hline
Cubic & 64.3 & 65.7 & 66.8 & 71.4 & 73.3 & 75.8 \\
Pillar & 62.4 & 64.1 & 65.2 & 69.4 & 71.5 & 73.2 \\
Sphere & \underline{64.8} & \underline{65.9} & \underline{67.2} & \underline{71.6} & \underline{73.8} & \underline{76.2} \\
Cylin. & \textbf{65.7} & \textbf{66.5} & \textbf{67.6} & \textbf{72.2} & \textbf{75.2} & \textbf{76.9}\\
\hline
\end{tabular}
\caption{\textbf{Backbone networks}. We test cubic, pillar, spherical, and cylindrical voxelization. All methods perform well.}
\label{table:ablation-backbone}
\end{subtable}
\hspace{\fill}
\begin{subtable}[t]{0.32\textwidth}
\setlength{\tabcolsep}{2pt}
\begin{tabular}{lcccccc}
\hline
\multirow{2}{*}{EMA ~~} & \multicolumn{3}{c}{SemanticKITTI} & \multicolumn{3}{c}{nuScenes} \\ \cmidrule(l){2-7}
& \makebox[1.5em][c]{10\%} & \makebox[1.5em][c]{20\%}  & \makebox[1.5em][c]{50\%} & \makebox[1.5em][c]{10\%}  & \makebox[1.5em][c]{20\%} & \makebox[1.5em][c]{50\%}  \\ 
\hline
0.5 & 65.2 & 65.9 & 67.1 & 71.5 & 74.5 & 76.1 \\
0.9 & 65.5 & 66.2 & 67.3 & 71.7 & 74.8 & 76.3 \\
0.99 & \textbf{65.7} & \textbf{66.5} & \textbf{67.6} & \textbf{72.2} & \textbf{75.2} & \textbf{76.9} \\
0.999 & 64.7 & 65.3 & 66.7 & 71.3 & 74.1 & 75.8 \\
\hline
\end{tabular}
\caption{\textbf{EMA ratio.} As the EMA ratio increases, segmentation performance shows a upward trend, reaching its peak at 0.99. }
\label{table:ablation-ema}
\end{subtable}
\hspace{\fill}
\begin{subtable}[t]{0.32\textwidth}
\setlength{\tabcolsep}{2pt}
\begin{tabular}{clccccc}
\hline
\# & Comp. & mIoU & Fog & Snow & Beam & Echo \\
\hline
\multirow{3}{*}{\rotatebox{90}{SK-C}} & CENet & 62.6 & 42.7 & 53.6 & 55.8 & 53.4 \\
& FRNet & \textbf{68.7} & 47.6 & 57.1 & \textbf{62.5} & \textbf{58.1} \\ 
& Ours & 67.8 & \textbf{56.2} & \textbf{58.0} & 58.5 & 57.9 \\ 
\hline
\multirow{3}{*}{\rotatebox{90}{NS-C}} & CENet & 73.3 & 67.0 & 61.6 & 50.0 & 53.3 \\
& FRNet & \textbf{79.0} & \textbf{69.1} & 69.5 & \textbf{68.3} & 58.7 \\ 
& Ours & 77.3 & 68.3 & \textbf{70.2} & 65.7 & \textbf{61.9} \\ 
\hline
\end{tabular}
\caption{\textbf{Robustness}.  \sysName~has comparable robustness with SoTA fully-supervised models in ``Fog'', ``Snow'', ``Wet'', and ``Echo'' conditions. }
\label{table:ablation-robust}
\end{subtable}
\label{table:ablation}
\vspace{-5pt}
\end{table*}

\noindent\textbf{Class-wise Performance. }
In terms of class-wise performance, as shown in Table \ref{tab:02}, our method performs well on the class ``parking'' but shows weaker performance on the class ``terrain''. This highlights \sysName's capability to identify distant objects effectively. Additionally, all methods demonstrate relatively low performance on the ``other-ground'', ``bicycle'', and ``traffic-sign'', which remains a significant challenge in semi-supervised LiDAR point cloud segmentation. Future strategies could focus on improving recognition for these classes.

\noindent\textbf{Segmentation Visualization. }
We further present typical examples of LiDAR point cloud semantic segmentation results at a 50\% annotation ratio in Fig.~\ref{fig04}, with orange boxes highlighting areas of significant differences between \sysName~and LaserMix~\cite{c17:kong2023}. Compared to LaserMix, \sysName~shows a distinct advantage in accurately segmenting distant areas with objects rapidly changing classes or shapes. This supports the design rationale behind our high and low temporal sensitivity flows, confirming the effectiveness of targeting regions with varying temporal dynamics.

\subsection{Ablation Study}

\noindent\textbf{(a) \sysName~Embedding Unit (HEU). }
HEU is the primary innovation of this work, focusing on both nearby regions with minimal class and shape changes and distant regions with more substantial variations. It mainly consists of the high temporal sensitivity flow (HTSF) and low temporal sensitivity flow (LTSF). Table~\ref{table:ablation}\subref{table:ablation-heu} presents a performance comparison for segmentation by showing results with HEU entirely disabled, HTSF-only, LTSF-only, and with the complete HEU module. It shows that  HTSF and LTSF significantly improve LiDAR segmentation performance. Furthermore, HTSF, which targets temporally sensitive regions with variations in both class and shape, contributes more prominently to performance enhancement than LTSF. The best results are achieved when both are combined in the HEU module.

\noindent\textbf{(b) Cross-Attention and Fusion. }
As shown in Fig.\ref{fig03}, when performing cross-attention between HTSF and LTSF, HTSF is used as the key (K) and value (V), while LTSF is used as the query (Q). We also test by using HTSF as Q and LTSF as Q, as well as simple addition or concatenation of HTSF and LTSF. The experimental results, reported in Table~\ref{table:ablation}\subref{table:ablation-fusion}, show that performing cross-attention between high-sensitive flow and low-sensitive flow improves performance by approximately 2\%. This indicates that the cross-attention between high and low temporal sensitivity features allows the model to better capture both dynamic and stable object information, leading to more accurate performance.

\noindent\textbf{(c) Multi-Voxel Aggregation. }
As described in Sec.~\ref{sec:multi-voxel-aggregation}, we apply multi-voxel aggregation among nearby voxels to reduce computation complexity in \sysName~Embedding Unit. We also conduct two additional experiments to reduce the number of input voxels. The first method randomly selects $m$ voxels, while the second method chooses the top-$m$ voxels with the highest point cloud density. Since both methods inevitably discard information from the unselected voxels, our proposed method outperforms these two approaches by 2-3\%, as reported in Table~\ref{table:ablation}\subref{table:ablation-MVA}.

\noindent\textbf{(d) Backbone Networks. }
In the voxelization step shown in Fig.~\ref{fig02}, we can arrange LiDAR point clouds in cubic~\cite{c10:zhou2018}, pillar~\cite{c30:lang2019}, and spherical~\cite{c31:lai2023spherical} formats, and leverage corresponding backbone networks to generate feature maps. As shown in Table~\ref{table:ablation}\subref{table:ablation-backbone}, all these voxelization methods perform well, indicating that \sysName~can effectively capture LiDAR spatial characteristics at both near and far ranges if the arrangement of point clouds inherently encodes distance-related properties. We choose the cylindrical voxelization in \sysName~since it performs the best.

\noindent\textbf{(e) EMA Ratio. }
The EMA update ratio is a relatively sensitive factor that impacts segmentation performance in previous semi-supervised mean-teacher architecture~\cite{c17:kong2023, c33:li2023}. 
In contrast, as shown in Table~\ref{table:ablation}\subref{table:ablation-ema}, changes in the update ratio have less impact on the performance outcomes, indicating \sysName~exhibits a high level of stability and robustness with respect to EMA ratio.

\noindent\textbf{(f) Performance on Out-of-distribution Datasets. }
In addition to conventional segmentation evaluation, we further investigate \sysName's~robustness under various data perturbations using SemanticKITTI-C and nuScenes-C datasets~\cite{yan2024benchmarking, kong2023robo3d}, and compare it with CENet~\cite{cheng2022cenet} and FRNet~\cite{xu2025frnet}. As shown in Table~\ref{table:ablation}\subref{table:ablation-robust}, \sysName~ exhibits comparable robustness under three types of perturbations: severe weather conditions (Fog, Snow), external disturbances (Beam-missing), and internal sensor failures (Echo). These results demonstrate the effectiveness of high-low temporal sensitive representation learning in various conditions.

\section{Conclusion and Limitation}\label{sec:conclusion}

In this paper, we propose a novel semi-supervised LiDAR point cloud segmentation method, \sysName, which effectively leverages temporal dynamics through the High Temporal Sensitivity Flow and Low Temporal Sensitivity Flow. By focusing on different regions with varying temporal characteristics, \sysName~significantly improves LiDAR semantic segmentation performance, especially for distant objects with rapidly changing shapes and categories. Our experimental results, evaluated on the widely used SemanticKITTI and nuScenes datasets, demonstrate that \sysName~outperforms both fully-supervised and state-of-the-art semi-supervised methods under limited annotations.

Since \sysName~is specifically designed for the autonomous driving domain, where it leverages point clouds from consecutive frames, it is not suitable for general object point cloud segmentation task, such as posed in PointNet~\cite{c38:qi2017} and KPConv~\cite{c29:thomas2019}.

\noindent \textbf{Acknowledgements:}  This work was supported by the National Natural Science Foundation of China under grant 62102307, U21A20462,
62372400, and 62372365, and Fundamental Research Funds for the Central Universities.

{
    \small
    \bibliographystyle{ieeenat_fullname}
    \bibliography{reference}

\begin{thebibliography}{46}
\providecommand{\natexlab}[1]{#1}
\providecommand{\url}[1]{\texttt{#1}}
\expandafter\ifx\csname urlstyle\endcsname\relax
  \providecommand{\doi}[1]{doi: #1}\else
  \providecommand{\doi}{doi: \begingroup \urlstyle{rm}\Url}\fi

\bibitem[Ando et~al.(2023)Ando, Gidaris, Bursuc, Puy, Boulch, and Marlet]{c40:ando2023}
Angelika Ando, Spyros Gidaris, Andrei Bursuc, Gilles Puy, Alexandre Boulch, and Renaud Marlet.
\newblock Rangevit: Towards vision transformers for 3d semantic segmentation in autonomous driving.
\newblock In \emph{Proceedings of the IEEE/CVF conference on computer vision and pattern recognition}, pages 5240--5250, 2023.

\bibitem[Ayg\"{u}n et~al.(2021)Ayg\"{u}n, Osep, Weber, Maximov, Stachniss, Behley, and Leal-Taix{\'e}]{c21:aygun2021}
Mehmet Ayg\"{u}n, Aljosa Osep, Mark Weber, Maxim Maximov, Cyrill Stachniss, Jens Behley, and Laura Leal-Taix{\'e}.
\newblock 4d panoptic lidar segmentation.
\newblock In \emph{Proceedings of the IEEE/CVF Conference on Computer Vision and Pattern Recognition}, pages 5527--5537, 2021.

\bibitem[Behley et~al.(2019)Behley, Garbade, Milioto, Quenzel, Behnke, Stachniss, and Gall]{c26:behley2019}
Jens Behley, Martin Garbade, Andres Milioto, Jan Quenzel, Sven Behnke, Cyrill Stachniss, and Jurgen Gall.
\newblock Semantickitti: A dataset for semantic scene understanding of lidar sequences.
\newblock In \emph{Proceedings of the IEEE/CVF international conference on computer vision}, pages 9297--9307, 2019.

\bibitem[Caesar et~al.(2020)Caesar, Bankiti, Lang, Vora, Liong, Xu, Krishnan, Pan, Baldan, and Beijbom]{c27:hulger2020}
Holger Caesar, Varun Bankiti, Alex~H. Lang, Sourabh Vora, Venice~Erin Liong, Qiang Xu, Anush Krishnan, Yu Pan, Giancarlo Baldan, and Oscar Beijbom.
\newblock nuscenes: A multimodal dataset for autonomous driving.
\newblock In \emph{Proceedings of the IEEE/CVF conference on computer vision and pattern recognition}, 2020.

\bibitem[Chang et~al.(2025)Chang, Xue, Liu, Pan, and Wei]{chang2025drivingrulesbenchmarkintegrating}
Xinyuan Chang, Maixuan Xue, Xinran Liu, Zheng Pan, and Xing Wei.
\newblock Driving by the rules: A benchmark for integrating traffic sign regulations into vectorized hd map.
\newblock In \emph{Proceedings of the IEEE/CVF Conference on Computer Vision and Pattern Recognition}, 2025.

\bibitem[Chen et~al.(2024)Chen, Tan, Zhang, Qu, and Xie]{c46:chen2024}
Yujun Chen, Xin Tan, Zhizhong Zhang, Yanyun Qu, and Yuan Xie.
\newblock Beyond the label itself: Latent labels enhance semi-supervised point cloud panoptic segmentation.
\newblock In \emph{Proceedings of the AAAI Conference on Artificial Intelligence}, pages 1245--1253, 2024.

\bibitem[Cheng et~al.(2022)Cheng, Han, and Xiao]{cheng2022cenet}
Hui-Xian Cheng, Xian-Feng Han, and Guo-Qiang Xiao.
\newblock Cenet: Toward concise and efficient lidar semantic segmentation for autonomous driving.
\newblock In \emph{2022 IEEE international conference on multimedia and expo (ICME)}, pages 01--06. IEEE, 2022.

\bibitem[Cheng et~al.(2021)Cheng, Hui, Xie, and Yang]{c34:cheng2021}
Mingmei Cheng, Le Hui, Jin Xie, and Jian Yang.
\newblock Sspc-net: Semi-supervised semantic 3d point cloud segmentation network.
\newblock In \emph{Proceedings of the AAAI conference on artificial intelligence}, pages 1140--1147, 2021.

\bibitem[Choy et~al.(2019)Choy, Gwak, and Savarese]{c22:choy2019}
Christopher Choy, JunYoung Gwak, and Silvio Savarese.
\newblock 4d spatio-temporal convnets: Minkowski convolutional neural networks.
\newblock In \emph{Proceedings of the IEEE/CVF conference on computer vision and pattern recognition}, pages 3075--3084, 2019.

\bibitem[Dosovitskiy et~al.(2021)Dosovitskiy, Beyer, Kolesnikov, Weissenborn, Zhai, Unterthiner, Dehghani, Minderer, Heigold, Gelly, Uszkoreit, and Houlsby]{c42:dosovitskiy2021}
Alexey Dosovitskiy, Lucas Beyer, Alexander Kolesnikov, Dirk Weissenborn, Xiaohua Zhai, Thomas Unterthiner, Mostafa Dehghani, Matthias Minderer, Georg Heigold, Sylvain Gelly, Jakob Uszkoreit, and Neil Houlsby.
\newblock An image is worth 16x16 words: Transformers for image recognition at scale.
\newblock In \emph{International Conference on Learning Representations}, 2021.

\bibitem[Fei et~al.(2021)Fei, Peng, Heidenreich, Bieder, and Stiller]{c56:fei2021pillarsegnet}
Juncong Fei, Kunyu Peng, Philipp Heidenreich, Frank Bieder, and Christoph Stiller.
\newblock Pillarsegnet: Pillar-based semantic grid map estimation using sparse lidar data.
\newblock In \emph{2021 IEEE intelligent vehicles symposium (IV)}, pages 838--844. IEEE, 2021.

\bibitem[He et~al.(2016)He, Zhang, Ren, and Sun]{c41:he2016}
Kaiming He, Xiangyu Zhang, Shaoqing Ren, and Jian Sun.
\newblock Deep residual learning for image recognition.
\newblock In \emph{Proceedings of the IEEE conference on computer vision and pattern recognition}, pages 770--778, 2016.

\bibitem[Huang et~al.(2021)Huang, Xie, Zhu, and Zhu]{c9:huang2021}
Siyuan Huang, Yichen Xie, Song-Chun Zhu, and Yixin Zhu.
\newblock Spatio-temporal self-supervised representation learning for 3d point clouds.
\newblock In \emph{Proceedings of the IEEE/CVF International Conference on Computer Vision}, pages 6535--6545, 2021.

\bibitem[Jhaldiyal and Chaudhary(2023)]{c8:jhaldiyal2023}
Alok Jhaldiyal and Navendu Chaudhary.
\newblock Semantic segmentation of 3d lidar data using deep learning: a review of projection-based methods.
\newblock \emph{Applied Intelligence}, 53\penalty0 (6):\penalty0 6844--6855, 2023.

\bibitem[Jiang et~al.(2021)Jiang, Shi, Tian, Lai, Liu, Fu, and Jia]{c13:jiang2021}
Li Jiang, Shaoshuai Shi, Zhuotao Tian, Xin Lai, Shu Liu, Chi-Wing Fu, and Jiaya Jia.
\newblock Guided point contrastive learning for semi-supervised point cloud semantic segmentation.
\newblock In \emph{Proceedings of the IEEE/CVF international conference on computer vision}, pages 6423--6432, 2021.

\bibitem[Kong et~al.(2023{\natexlab{a}})Kong, Liu, Chen, Ma, Zhu, Li, Hou, Qiao, and Liu]{c54:kong2023rethinking}
Lingdong Kong, Youquan Liu, Runnan Chen, Yuexin Ma, Xinge Zhu, Yikang Li, Yuenan Hou, Yu Qiao, and Ziwei Liu.
\newblock Rethinking range view representation for lidar segmentation.
\newblock In \emph{Proceedings of the IEEE/CVF International Conference on Computer Vision}, pages 228--240, 2023{\natexlab{a}}.

\bibitem[Kong et~al.(2023{\natexlab{b}})Kong, Liu, Li, Chen, Zhang, Ren, Pan, Chen, and Liu]{kong2023robo3d}
Lingdong Kong, Youquan Liu, Xin Li, Runnan Chen, Wenwei Zhang, Jiawei Ren, Liang Pan, Kai Chen, and Ziwei Liu.
\newblock Robo3d: Towards robust and reliable 3d perception against corruptions.
\newblock \emph{arXiv preprint arXiv:2303.17597}, 2023{\natexlab{b}}.

\bibitem[Kong et~al.(2023{\natexlab{c}})Kong, Ren, Pan, and Liu]{c17:kong2023}
Lingdong Kong, Jiawei Ren, Liang Pan, and Ziwei Liu.
\newblock Lasermix for semi-supervised lidar semantic segmentation.
\newblock In \emph{Proceedings of the IEEE/CVF Conference on Computer Vision and Pattern Recognition}, pages 21705--21715, 2023{\natexlab{c}}.

\bibitem[Kong et~al.(2024)Kong, Xu, Ren, Zhang, Pan, Chen, Ooi, and Liu]{c53:kong2024multi}
Lingdong Kong, Xiang Xu, Jiawei Ren, Wenwei Zhang, Liang Pan, Kai Chen, Wei~Tsang Ooi, and Ziwei Liu.
\newblock Multi-modal data-efficient 3d scene understanding for autonomous driving.
\newblock \emph{arXiv preprint arXiv:2405.05258}, 2024.

\bibitem[Lai et~al.(2023)Lai, Chen, Lu, Liu, and Jia]{c31:lai2023spherical}
Xin Lai, Yukang Chen, Fanbin Lu, Jianhui Liu, and Jiaya Jia.
\newblock Spherical transformer for lidar-based 3d recognition.
\newblock In \emph{Proceedings of the IEEE/CVF Conference on Computer Vision and Pattern Recognition}, pages 17545--17555, 2023.

\bibitem[Lan et~al.(2024)Lan, Wang, Xia, Nai, Nie, Ding, and Han]{lan2024bullydetect}
Bo Lan, Fei Wang, Lekun Xia, Fan Nai, Shiqiang Nie, Han Ding, and Jinsong Han.
\newblock Bullydetect: Detecting school physical bullying with wi-fi and deep wavelet transformer.
\newblock \emph{IEEE Internet of Things Journal}, 2024.

\bibitem[Lang et~al.(2019)Lang, Vora, Caesar, Zhou, Yang, and Beijbom]{c30:lang2019}
Alex~H Lang, Sourabh Vora, Holger Caesar, Lubing Zhou, Jiong Yang, and Oscar Beijbom.
\newblock Pointpillars: Fast encoders for object detection from point clouds.
\newblock In \emph{Proceedings of the IEEE/CVF conference on computer vision and pattern recognition}, pages 12697--12705, 2019.

\bibitem[Li and Dong(2024)]{c48:li2024}
Jianan Li and Qiulei Dong.
\newblock Density-guided semi-supervised 3d semantic segmentation with dual-space hardness sampling.
\newblock In \emph{Proceedings of the IEEE/CVF Conference on Computer Vision and Pattern Recognition}, pages 3260--3269, 2024.

\bibitem[Li et~al.(2023{\natexlab{a}})Li, Dai, Han, and Ding]{c32:li2023}
Jiale Li, Hang Dai, Hao Han, and Yong Ding.
\newblock Mseg3d: Multi-modal 3d semantic segmentation for autonomous driving.
\newblock In \emph{Proceedings of the IEEE/CVF conference on computer vision and pattern recognition}, pages 21694--21704, 2023{\natexlab{a}}.

\bibitem[Li et~al.(2023{\natexlab{b}})Li, Luo, and Yang]{c28:li2023}
Jinyu Li, Chenxu Luo, and Xiaodong Yang.
\newblock Pillarnext: Rethinking network designs for 3d object detection in lidar point clouds.
\newblock In \emph{Proceedings of the IEEE/CVF Conference on Computer Vision and Pattern Recognition}, pages 17567--17576, 2023{\natexlab{b}}.

\bibitem[Li et~al.(2023{\natexlab{c}})Li, Shum, and Breckon]{c33:li2023}
Li Li, Hubert~PH Shum, and Toby~P Breckon.
\newblock Less is more: Reducing task and model complexity for 3d point cloud semantic segmentation.
\newblock In \emph{Proceedings of the IEEE/CVF Conference on Computer Vision and Pattern Recognition}, pages 9361--9371, 2023{\natexlab{c}}.

\bibitem[Li et~al.(2022)Li, Xie, Shen, Ke, Qiao, Ren, Lin, and Ma]{c12:li2022}
Mengtian Li, Yuan Xie, Yunhang Shen, Bo Ke, Ruizhi Qiao, Bo Ren, Shaohui Lin, and Lizhuang Ma.
\newblock Hybridcr: Weakly-supervised 3d point cloud semantic segmentation via hybrid contrastive regularization.
\newblock In \emph{Proceedings of the IEEE/CVF conference on computer vision and pattern recognition}, pages 14930--14939, 2022.

\bibitem[Lin et~al.(2017)Lin, Goyal, Girshick, He, and Doll{\'a}r]{c51:lin2017}
Tsung-Yi Lin, Priya Goyal, Ross Girshick, Kaiming He, and Piotr Doll{\'a}r.
\newblock Focal loss for dense object detection.
\newblock In \emph{Proceedings of the IEEE international conference on computer vision}, pages 2980--2988, 2017.

\bibitem[Liu et~al.(2023)Liu, Chang, Liu, Wu, Ma, and Qi]{c37:liu2023}
Jiahui Liu, Chirui Chang, Jianhui Liu, Xiaoyang Wu, Lan Ma, and Xiaojuan Qi.
\newblock Mars3d: A plug-and-play motion-aware model for semantic segmentation on multi-scan 3d point clouds.
\newblock In \emph{Proceedings of the IEEE/CVF Conference on Computer Vision and Pattern Recognition}, pages 9372--9381, 2023.

\bibitem[Loshchilov and Hutter(2019)]{c49:loshchilov2018}
Ilya Loshchilov and Frank Hutter.
\newblock Decoupled weight decay regularization.
\newblock In \emph{International Conference on Learning Representations}, 2019.

\bibitem[Qi et~al.(2017)Qi, Su, Mo, and Guibas]{c38:qi2017}
Charles~R Qi, Hao Su, Kaichun Mo, and Leonidas~J Guibas.
\newblock Pointnet: Deep learning on point sets for 3d classification and segmentation.
\newblock In \emph{Proceedings of the IEEE conference on computer vision and pattern recognition}, pages 652--660, 2017.

\bibitem[Reichardt et~al.(2023)Reichardt, Ebert, and Wasenm{\"u}ller]{reichardt2023360deg}
Laurenz Reichardt, Nikolas Ebert, and Oliver Wasenm{\"u}ller.
\newblock 360deg from a single camera: a few-shot approach for lidar segmentation.
\newblock In \emph{Proceedings of the IEEE/CVF International Conference on Computer Vision}, pages 1075--1083, 2023.

\bibitem[Shi et~al.(2022)Shi, Wei, Li, Liu, and Lin]{c24:shi2022}
Hanyu Shi, Jiacheng Wei, Ruibo Li, Fayao Liu, and Guosheng Lin.
\newblock Weakly supervised segmentation on outdoor 4d point clouds with temporal matching and spatial graph propagation.
\newblock In \emph{Proceedings of the IEEE/CVF conference on computer vision and pattern recognition}, pages 11840--11849, 2022.

\bibitem[Shi et~al.(2023)Shi, Wang, Yu, Li, Hong, Wang, and Gong]{shi2023exploiting}
Jingang Shi, Yusi Wang, Zitong Yu, Guanxin Li, Xiaopeng Hong, Fei Wang, and Yihong Gong.
\newblock Exploiting multi-scale parallel self-attention and local variation via dual-branch transformer-cnn structure for face super-resolution.
\newblock \emph{IEEE Transactions on Multimedia}, 26:\penalty0 2608--2620, 2023.

\bibitem[Tarvainen and Valpola(2017)]{c14:tarvainen2017}
Antti Tarvainen and Harri Valpola.
\newblock Mean teachers are better role models: Weight-averaged consistency targets improve semi-supervised deep learning results.
\newblock \emph{Advances in neural information processing systems}, 30, 2017.

\bibitem[Thomas et~al.(2019)Thomas, Qi, Deschaud, Marcotegui, Goulette, and Guibas]{c29:thomas2019}
Hugues Thomas, Charles~R Qi, Jean-Emmanuel Deschaud, Beatriz Marcotegui, Fran{\c{c}}ois Goulette, and Leonidas~J Guibas.
\newblock Kpconv: Flexible and deformable convolution for point clouds.
\newblock In \emph{Proceedings of the IEEE/CVF international conference on computer vision}, pages 6411--6420, 2019.

\bibitem[Unal et~al.(2024)Unal, Dai, Hoyer, Can, and Van~Gool]{c47:unal2024}
Ozan Unal, Dengxin Dai, Lukas Hoyer, Yigit~Baran Can, and Luc Van~Gool.
\newblock 2d feature distillation for weakly-and semi-supervised 3d semantic segmentation.
\newblock In \emph{Proceedings of the IEEE/CVF Winter Conference on Applications of Computer Vision}, pages 7336--7345, 2024.

\bibitem[Vaswani et~al.(2017)Vaswani, Shazeer, Parmar, Uszkoreit, Jones, Gomez, Kaiser, and Polosukhin]{c25:vaswani2017}
Ashish Vaswani, Noam Shazeer, Niki Parmar, Jakob Uszkoreit, Llion Jones, Aidan~N Gomez, {\L}ukasz Kaiser, and Illia Polosukhin.
\newblock Attention is all you need.
\newblock \emph{Advances in neural information processing systems}, 30, 2017.

\bibitem[Xiao et~al.(2022)Xiao, Huang, Guan, Cui, Lu, and Shao]{xiao2022polarmix}
Aoran Xiao, Jiaxing Huang, Dayan Guan, Kaiwen Cui, Shijian Lu, and Ling Shao.
\newblock Polarmix: A general data augmentation technique for lidar point clouds.
\newblock \emph{Advances in Neural Information Processing Systems}, 35:\penalty0 11035--11048, 2022.

\bibitem[Xu et~al.(2025)Xu, Kong, Shuai, and Liu]{xu2025frnet}
Xiang Xu, Lingdong Kong, Hui Shuai, and Qingshan Liu.
\newblock Frnet: Frustum-range networks for scalable lidar segmentation.
\newblock \emph{IEEE Transactions on Image Processing}, 2025.

\bibitem[Xu et~al.(2023)Xu, Yuan, Zhao, Zhang, and Gao]{c35:xu2023}
Zongyi Xu, Bo Yuan, Shanshan Zhao, Qianni Zhang, and Xinbo Gao.
\newblock Hierarchical point-based active learning for semi-supervised point cloud semantic segmentation.
\newblock In \emph{Proceedings of the IEEE/CVF International Conference on Computer Vision}, pages 18098--18108, 2023.

\bibitem[Yan et~al.(2024{\natexlab{a}})Yan, Wang, Qian, Ding, Han, and Wei]{c43:yan2024}
Kangwei Yan, Fei Wang, Bo Qian, Han Ding, Jinsong Han, and Xing Wei.
\newblock Person-in-wifi 3d: End-to-end multi-person 3d pose estimation with wi-fi.
\newblock In \emph{Proceedings of the IEEE/CVF Conference on Computer Vision and Pattern Recognition}, pages 969--978, 2024{\natexlab{a}}.

\bibitem[Yan et~al.(2024{\natexlab{b}})Yan, Zheng, Xue, Li, Cui, and Dai]{yan2024benchmarking}
Xu Yan, Chaoda Zheng, Ying Xue, Zhen Li, Shuguang Cui, and Dengxin Dai.
\newblock Benchmarking the robustness of lidar semantic segmentation models.
\newblock \emph{International Journal of Computer Vision}, 132\penalty0 (7):\penalty0 2674--2697, 2024{\natexlab{b}}.

\bibitem[Ye et~al.(2023)Ye, Zhou, Chen, Xie, Wang, Wang, and Foroosh]{c55:ye2023lidarmultinet}
Dongqiangzi Ye, Zixiang Zhou, Weijia Chen, Yufei Xie, Yu Wang, Panqu Wang, and Hassan Foroosh.
\newblock Lidarmultinet: Towards a unified multi-task network for lidar perception.
\newblock In \emph{Proceedings of the AAAI Conference on Artificial Intelligence}, pages 3231--3240, 2023.

\bibitem[Zhou and Tuzel(2018)]{c10:zhou2018}
Yin Zhou and Oncel Tuzel.
\newblock Voxelnet: End-to-end learning for point cloud based 3d object detection.
\newblock In \emph{Proceedings of the IEEE conference on computer vision and pattern recognition}, pages 4490--4499, 2018.

\bibitem[Zhu et~al.(2021)Zhu, Zhou, Wang, Hong, Ma, Li, Li, and Lin]{c52:zhu2021cylindrical}
Xinge Zhu, Hui Zhou, Tai Wang, Fangzhou Hong, Yuexin Ma, Wei Li, Hongsheng Li, and Dahua Lin.
\newblock Cylindrical and asymmetrical 3d convolution networks for lidar segmentation.
\newblock In \emph{Proceedings of the IEEE/CVF conference on computer vision and pattern recognition}, pages 9939--9948, 2021.

\end{thebibliography}
}

\end{document}